\documentclass{article}

\PassOptionsToPackage{numbers}{natbib}
 
\usepackage[preprint]{nips_2018}

\usepackage[utf8]{inputenc} 
\usepackage[T1]{fontenc}    
\usepackage{hyperref}       
\usepackage{url}            
\usepackage{booktabs}       
\usepackage{amsfonts}       
\usepackage{nicefrac}       
\usepackage{microtype}      

\usepackage{float}
\usepackage{graphicx}
\usepackage{multirow}
\usepackage{amsmath}

\usepackage[usenames, dvipsnames]{color}

\usepackage{caption}
\usepackage{subcaption}

\makeatletter
\newcommand{\specificthanks}[1]{\@fnsymbol{#1}}
\makeatother

\title{Manifold regularization with GANs for semi-supervised learning}

\author{
Bruno Lecouat\thanks{All code and hyperparameters may be found at \url{https://github.com/bruno-31/gan-manifold-reg}}
  \textsuperscript{ , \specificthanks{2}} \\
Institute for Infocomm Research, A*STAR\\
\texttt{bruno\_lecouat@i2r.a-star.edu.sg}
\And
Chuan-Sheng Foo\thanks{Equal contribution}\label{yale-uni}\\
Institute for Infocomm Research, A*STAR\\
\texttt{foocs@i2r.a-star.edu.sg}
\And
Houssam Zenati\\
Institute for Infocomm Research, A*STAR\\
\texttt{houssam\_zenati@i2r.a-star.edu.sg}
\And
Vijay Chandrasekhar\\
Institute for Infocomm Research, A*STAR\\
\texttt{vijay@i2r.a-star.edu.sg} \\
}

\begin{document}

\maketitle

\begin{abstract}
Generative Adversarial Networks are powerful generative models that are able to model the manifold of natural images. We leverage this property to perform manifold regularization by approximating a variant of the Laplacian norm using a Monte Carlo approximation that is easily computed with the GAN. When incorporated into the semi-supervised feature-matching GAN we achieve state-of-the-art results for GAN-based semi-supervised learning on CIFAR-10 and SVHN benchmarks, with a method that is significantly easier to implement than competing methods. We also find that manifold regularization improves the quality of generated images, and is affected by the quality of the GAN used to approximate the regularizer.
\end{abstract}

\section{Introduction}

Deep neural network classifiers typically require large labeled datasets to obtain high predictive performance. Obtaining such a dataset could be time and cost prohibitive especially for applications where careful expert labeling is required, for instance, in healthcare and medicine. Semi-supervised learning algorithms that enable models to be learned from a small amount of labeled data augmented with (large amounts of) unlabeled data have the potential to vastly reduce this labeling burden. 

Fundamentally, semi-supervised learning requires assumptions relating the distribution of the data $\mathcal{P}_x$ (which can be derived from the unlabeled data) to the classification task \cite{ChapelleSemi-SupervisedBy}. For instance, the classic manifold regularization framework \cite{Belkin2006ManifoldExamples} for semi-supervised learning makes the assumption that that the data lie on a low-dimensional manifold $\mathcal{M}$ and moreover that a classifier $f$ is smooth on this manifold, so nearby points on the manifold are assigned similar labels. Algorithms based on this framework enforce a classifier's invariance to local perturbations on the manifold by penalizing its Laplacian norm $\left \| f \right \|_L^2 = \int _{x\in M} \left \| \nabla_\mathcal{M} f(x) \right \|^2 \mathrm{d} \mathcal{P}_X(x)$. More generally, regularization terms penalizing classifier gradients in regions of high data density have also been proposed \cite{BousquetMeasureRegularization}.

Recently, generative adversarial networks (GANs) have been used for semi-supervised learning, where they are competitive with state-of-the-art methods for semi-supervised image classification \cite{DaiGoodGAN, Kumar2017ImprovedInvariances, Qi2018GlobalNets}. GAN-based semi-supervised learning methods typically build upon the formulation in \cite{SalimansImprovedGANs}, where the discriminator is extended to determine the specific class of an image or whether it is generated; by contrast, the original GAN's discriminator is only expected to determine whether an image is real or generated. Another key application for GANs is image synthesis, where they have been shown to model the image manifold well \cite{Zhu2016GenerativeManifold}. Recent work \cite{Kumar2017ImprovedInvariances,Qi2018GlobalNets} has used this property of GANs to enforce discriminator invariance on the image manifold, resulting in improved accuracy on semi-supervised image classification. 

In this work, we leverage the ability of GANs to model the image manifold to efficiently approximate the Laplacian norm and related regularization terms through Monte-Carlo integration. We show that classifiers (with varying network architectures) regularized with our method outperform baselines on the SVHN and CIFAR-10 benchmark datasets. In particular, when applied to the semi-supervised feature-matching GAN \cite{SalimansImprovedGANs}, our method achieves state-of-the-art performance amongst GAN-based methods, 
and is highly competitive with other non-GAN approaches especially when the number of labeled examples is small. We show that manifold regularization improves the quality of generated images as measured by Inception and FID scores when applied to the semi-supervised feature-matching GAN, thus linking manifold regularization to recent work on gradient penalties for stabilizing GAN training \cite{Gulrajani2017ImprovedGANs, Mescheder2017TheGANs}. We also found that generator quality (as measured by the quality of generated images) influences the benefit provided by our manifold regularization strategy in that using a better quality generator results in larger improvements in classification performance over a supervised baseline.


\section{Related Work}

There have been several works adapting GANs for semi-supervised learning. One approach is to change the standard binary discriminator of a standard GAN to one that predicts class labels of labelled examples, while enforcing the constraint that generated data should result in uncertain classifier predictions \cite{catgan}. The related approach of \cite{SalimansImprovedGANs} also uses the discriminator of the GAN as the final classifier, but instead modifies it to predict K+1 probabilities (K real classes and the generated class). This approach was shown to work well when combined with a feature matching loss for the generator. The work of \cite{Li2017TripleNets} introduces an additional classifier as well as uses a conditional generator instead of adapting the discriminator to overcome limitations with the two-player formulation of standard GANs in the context of semi-supervised learning.

The idea of encouraging local invariances dates back to the TangentProp algorithm \cite{Simard1991TangentNetwork} where manifold gradients at input data points are estimated using explicit transformations of the data that keep it on the manifold, for example small rotations and translations. Since then other approaches have tried to estimate these tangent directions in different ways. High order contractive autoencoders were used in \cite{Rifai2011ContractiveExtraction} to capture the structure of the manifold; this representation learning algorithm was then used to encourage a classifier to be insensitive to local direction changes along the manifold \cite{Rifai2011TheClassifier}. 
This approach was recently revisited in the context of GANs \cite{Kumar2017ImprovedInvariances}, where the tangent space to the data manifold is estimated using GANs with an encoder in order to inject invariance into the classifier. In addition, this work also explored the use of an additional ambient regularization term which promotes invariance of the discriminator to perturbations on training images along all directions in the data space. The proposed method is competitive with the state-of-the-art GAN method of \cite{DaiGoodGAN}, which argues that a generator that generates images that are in the complement of the training data distribution is necessary for good semi-supervised learning performance. Most recently \cite{Qi2018GlobalNets} proposed the use of a local GAN which attempts to model the local manifold geometry around data points without the need for an encoder. The local GAN is then used to approximate the Laplacian norm for semi-supervised learning, and is shown to enable state-of-the-art classification results. 

Aside from GAN-based approaches, Virtual Adversarial Training (VAT) \cite{Miyato2017VirtualLearning}, which is based on constraining model predictions to be consistent to local perturbation, has also achieved state-of-the-art performance on benchmarks. Specifically, VAT smooths predictions of the classifier over adversarial examples centered around labeled and unlabeled examples. 

Other recent works are based on the self-training paradigm \cite{ChapelleSemi-SupervisedBy}. Such methods label the unlabelled data using classifiers trained on the labelled data, and then use this expanded labelled dataset for training a final classifier. Recent progress has resulted from clever use of ensembling to produce better predictions on the unlabelled data. For instance, instead of simply using predictions of a model trained on the labelled data, \cite{Laine2017TemporalLearning} ensembled the predictions of the model under different perturbations or at different time steps. In follow-up work, the Mean Teacher method \cite{TarvainenMeanResults} averages model weights (instead of predictions) at different time steps using exponential moving averages, and achieved state-of-the-art performance on benchmark image datasets.


\section{Manifold regularization with GANs}

We present an approach to approximate any density-based regularization term of the form 
$\Omega(f) = \int_{x\in \mathcal{M}} L(f) \mathrm{d} {\chi(\mathcal{P}_X)} $ 
\cite{BousquetMeasureRegularization} with $L$ denoting a measure of smoothness of the classifier function $f$ and $\chi$ a strictly-increasing function. Such regularizers enforce classifier smoothness in regions of high data density. This class of regularizers includes the Laplacian norm with $L(f) = \left\|\nabla_\mathcal{M} f \right\|^2$ and $\chi$ the identity function. We focus on the following variant of the Laplacian norm
\[
\Omega(f) = \int_{x\in \mathcal{M}} \left\|\nabla_\mathcal{M} f \right\|_F \mathrm{d} \mathcal{P}_X
\]
with $L(f) = \left\|\nabla_\mathcal{M} f \right\|_F$ in this work and show how it can be approximated efficiently \footnote{In our early experiments we used the regular Laplacian norm but found that this variant worked better in a wider range of settings; results using the Laplacian norm are included in the Supplementary Material.}. Our approach relies on two commonly held assumptions about GANs:
\begin{enumerate}
\item GANs are able to model the distribution over images \cite{Radford2016UnsupervisedNetworks}, such that samples from the GAN are distributed approximately as $\mathcal{P}_X(x)$, the marginal distribution over images $x$.
\item GANs learn the image manifold \cite{Radford2016UnsupervisedNetworks,Zhu2016GenerativeManifold}. Specifically, we assume that the generator $g$ learns a mapping from the low-dimensional latent space with coordinates $z$ to the image manifold embedded in a higher-dimensional space, enabling us to compute gradients on the manifold by taking derivatives with respect to $z$ \cite{Kumar2017ImprovedInvariances,Shao2017TheModels} .
\end{enumerate}
With these assumptions, we may approximate $\Omega(f)$ as follows, where we list the relevant assumption above each approximation step
\[
   \Omega(f) = 
   \int_{x\in \mathcal{M}} \left\|\nabla_\mathcal{M} f \right\|_F \mathrm{d} \mathcal{P}
   \overset{\mathrm{(1)}}{\approx} \frac{1}{n} \sum_{i=1}^{n} \left \| \nabla_\mathcal{M} f(g(z^{(i)})) \right \|_F
   \overset{\mathrm{(2)}}{\approx} \frac{1}{n} \sum_{i=1}^{n} \left \| J_z f(g(z^{(i)})) \right \|_F.
\]
Here, $J_z$ denotes the Jacobian matrix of partial derivatives of classifier outputs $f$ with respect to latent generator variables $z$ \footnote{In our experiments, we defined $f$ to be the logits of the softmax output layer instead of the resultant normalized probabilities as we found it gave better performance. 
}. 
Computing gradients of $\Omega(f)$ during model learning is computationally prohibitive for deep neural networks as it requires computing the Hessian of a model with large numbers of parameters \footnote{In fact, for multi-class classifiers $f$, we need to compute a Hessian \emph{tensor} -- one matrix for each component (class output) of $f$, which quickly becomes impractical even with moderate numbers of classes.}. We hence used stochastic finite differences to approximate the gradient term for computational efficiency. 

\begin{figure}
\centering
\begin{subfigure}[b]{.5\textwidth}
  \raggedright
  \includegraphics[width=1.2\linewidth]{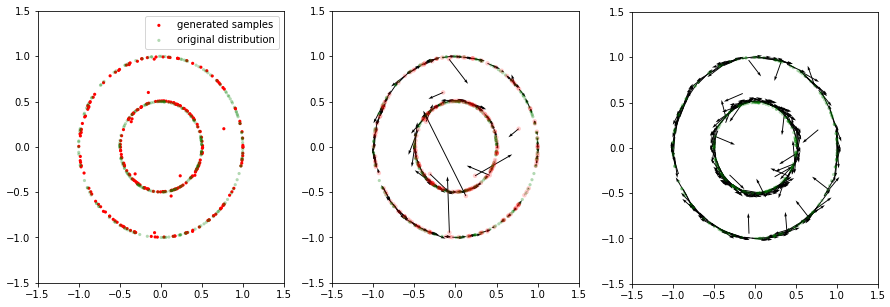}
  \caption{}
  \label{fig:sub1}
\end{subfigure}%
\begin{subfigure}[b]{0.5\textwidth}
  \raggedleft
  \includegraphics[width=.7\linewidth]{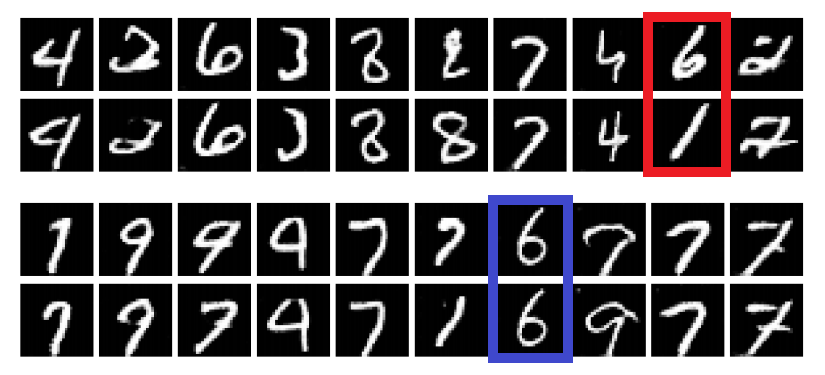}
  \caption{}
  \label{fig:sub2}
\end{subfigure}
  \caption{Issues with GAN-derived manifold gradients. (a) Left: Samples and generated samples from a GAN trained on the two circles dataset. Middle: Manifold gradients from the trained GAN. Right: Manifold gradients normalized to unit norm. 
The GANs approximate the data manifold well for this toy example and normalizing gradients mitigates issues with noisy gradients.
(b) Effect of perturbations on the latent code of a GAN trained on MNIST. Each pair of rows shows the generated example from the latent code (top) and the generated example from the perturbed latent code (bottom). Random perturbations of equal norm in the latent space can have almost no effect (blue box) or a large effect (red box) on generated examples.}
  \label{gradient}
\end{figure}

To motivate the specific approximation we used, we first illustrate several issues with manifold gradients as estimated with a GAN when considering the obvious candidate approximation $\left\| J_z f(g(z^{(i)})) \right \|_F \approx \left\|f(g(z^{(i)})) - f(g(z^{(i)} + \delta))\right\|_F, \delta \sim \mathcal{N}(0, \sigma^2 I)$ (Figure \ref{gradient}), using the Two Circles dataset and MNIST. The Two Circles dataset is an example of data lying on disjoint manifolds. In this case, even though the GAN is able to accurately model the data distribution (Figure \ref{gradient}a, left), we see several instances where the manifold gradients as per the GAN are extremely noisy (Figure \ref{gradient}a, center) with large magnitudes. If data from the inner and outer circle were to belong to different classes, enforcing classifier smoothness at those points in the manifold with these large noisy gradients could result in the classifier predicting similar values for both circles (as $g(z^{(i)})$ and $g(z^{(i)} + \delta) \approx g(z^{(i)}) + J_z(g(z^{(i)})) \delta$ lie on different circles), causing erroneous classifications. At the other extreme, there are points on the manifold where the estimated gradient has such small magnitude such that the regularizer has minimal smoothing effect. These issues are also evident on the MNIST dataset (Figure \ref{gradient}b), where we directly show how $g(z^{(i)})$ and $g(z^{(i)} + \delta)$ can lie on different manifolds (red rectangle) or be virtually identical (blue rectangle), resulting in over-smoothing and under-smoothing respectively of the classifier function.

In light of these issues arising from the magnitude of manifold gradients, we used the following approximation that takes a step of tunable size $\epsilon$ in the direction of the manifold gradient, thus ignoring the magnitude of the gradient while enforcing smoothness in its direction
\[  \Omega(f) \approx \frac{1}{n} \sum_{i=1}^{n}  \left\|f\left(g \left( z^{(i)} \right) \right)-f\left (  g\left (z^{(i)}\right )+\epsilon \bar{r}\left(z^{(i)}\right)\right)\right \|_F.
\]
Here $r(z) = g(z+\eta \: \bar{\delta}) - g(z^{(i)}), \delta \sim \mathcal{N}(0,I)$ is an approximation of the manifold gradient at $z$ with tunable step size $\eta$, and $\bar{v} =\frac{v}{\left \| v \right \|}$ denotes a unit vector.

We wish to highlight that our approach only relies on training a standard GAN. In contrast to the approach of \cite{Kumar2017ImprovedInvariances}, we do not explicitly enforce classifier smoothness on input data points, allowing us to avoid the added complexity of learning an encoder network to determine the underlying manifold coordinates for a data sample, as well as other tricks required to estimate tangent directions. The concurrent work \cite{Qi2018GlobalNets} develops an alternate and elegant solution to the issues we identified by learning a local GAN instead, but at the price of training a GAN with a more complex local generator.

\section{Experiments}

\subsection{Experimental setup}
Our experimental setup follows \cite{TarvainenMeanResults} -- we separated 10\% of the training data on CIFAR-10 into a validation set, and did similarly for SVHN. We designed our algorithms and chose our hyperparameters based on this validation set. 
We perform semi-supervised training using a small fraction of the labeled training data containing an equal number of examples from each class; the remaining training images are used as unlabeled data. 
Note that the classifier we used for evaluation had weights that were the exponential moving average of classifier weights obtained during training (this technique was also used in \cite{SalimansImprovedGANs}). We report the error rates on the test set for models which performed best on the validation set.  
Details of hyperparameters and network architectures we used in our experiments can be found in the Supplementary Material.

\subsection{Incorporating manifold regularization into semi-supervised GANs}
We first evaluated our regularization method when incorporated into the semi-supervised GAN framework. We reproduced the semi-supervised feature-matching GAN of \citep{SalimansImprovedGANs} and added our manifold regularizer to the model. The final loss function of the discriminator is:
\[L= L_{supervised} + L_{unsupervised}+ \gamma_m \Omega_{manifold} \text{  ,where} \] 
\begin{align*} 
\label{ldis}
\Omega_{manifold} &=  \: \mathbb{E}_{z \sim U(z), \delta \sim N(\delta))} \left\|f\left(g \left( z \right) \right)-f\left (  g\left (z\right )+\epsilon \bar{r}\right)\right \|_2\\
L_{supervised} &= - \mathbb{E}_{x,y \sim p_{data}(x,y)} \left[ \log{p_{f} (y|x,y<K+1)} \right]  \\
L_{unsupervised} &= -   \mathbb{E}_{x \sim p_{data}(x)} \left[ \log{[1- p_{f} (y=K+1|x)]} \right]  - \mathbb{E}_{x \sim g} \left[ \log {[p_{f} (y=K+1|x)]} \right] \\
\end{align*}
and we used the feature matching loss for our generator, $ \left\| \mathbb{E}_{x \sim p_{data}} h(x)		-    \mathbb{E}_{z \sim p_{z}(z)} h(g(z))\right\|$. Here, $h(x)$ denotes activations on an intermediate layer of the discriminator. We also checked if the anisotropic regularization along manifold directions in the data space that we used provides additional benefits over simple ambient regularization in the data space. Specifically, we evaluated the ambient regularizer $\lambda \mathbb{E}_{x\sim p_d(x)} \left \|       J_x f        \right \| $, as proposed in \cite{Kumar2017ImprovedInvariances} that we similarly approximate using a stochastic finite difference as shown below: $L = L_{supervised}+L_{unsupervised}+\gamma_a \Omega_{ambient}$ , where
\[
\Omega_{ambient}=\mathbb{E}_{\delta \sim N(\delta))} \left\|f\left(x \right)-f\left (x+\epsilon \: \bar{\delta}\right) \right \|_F.
\]

\begin{table}[h]
\centering
\caption{Error rate on CIFAR-10 average over 4 runs with different random seeds. Results were obtained without data augmentation.}
\label{cifar10:sslgan}
\begin{tabular}{lllll}
                          										 									& 1000 labels (2\%)  &2000 labels (4\%)        & 4000 labels  (8\%)          \\
  \textbf{CIFAR-10}         																			 & 50000 images                   &50000 images   & 50000 images                    \\ \hline

$\Pi$ model \cite{Laine2017TemporalLearning}						&     &    &       16.55 $\pm$ 0.29                      \\
Mean Teacher \cite{TarvainenMeanResults}		                 	&   30.62 $\pm$1.13   & 23.14 $\pm$ 0.46   &       17.74 $\pm$ 0.30                      \\  

VAT (large) \cite{Miyato2017VirtualLearning}                     &                                    & & 14.18                      \\
VAT+EntMin(Large)\cite{Miyato2017VirtualLearning}       &                                   &   & 13.15                        \\ \hline
Improved GAN  \cite{SalimansImprovedGANs}      	             & 21.83 $\pm$ 2.01           &19.61 $\pm$ 2.09    & 18.63 $\pm$ 2.32            \\
Improved Semi-GAN\cite{Kumar2017ImprovedInvariances} &  19.52 $\pm$1.5    &    & 16.20 $\pm$ 1.6                \\
ALI \cite{Dumoulin2017AdversariallyInference}                    & 19.98 $\pm$ 0.89           &19.09 $\pm$ 0.44   & 17.99 $\pm$ 1.62  \\
Triple-GAN \cite{Li2017TripleNets}                                              &									&							&16.99 $\pm$ 0.36\\
Bad GAN   \cite{DaiGoodGAN}                            			             &                                   &                            & 14.41 $\pm$ 0.30              \\
Local GAN \cite{Qi2018GlobalNets}													& 17.44 $\pm$ 0.25       &&                   14.23 $\pm$ 0.27  \\   \hline
Improved GAN (ours)                                                                    &17.50 $\pm$ 0.34       &    16.80 $\pm$ 0.07   & 15.5 $\pm$ 0.35    \\
Ambient regularization (ours)                                    	&16.81 $\pm$ 0.21                       & 15.99 $\pm$ 0.14                                             &  14.75 $\pm$ 0.37                         \\
\textbf{Manifold  regularization (ours)}                                                & \textbf{16.37 $\pm$ 0.42}  & \textbf{15.25 $\pm$ 0.35}                             &\textbf{14.34 $\pm$ 0.17}                  \\ \hline
\end{tabular}
\end{table}

\begin{table}[h]
\centering
\caption{Error rate on SVHN average over 4 runs with different random seeds. Results were obtained without data augmentation.}
\label{svhn:sslgan}
\begin{tabular}{lllll}
                                                                & 500 labels(0.3\%)       & 1000 labels(1.4\%)    \\
 \textbf{SVHN}                                       							 & 73257 images & 73257 images \\ \hline
$\Pi$ model \cite{Laine2017TemporalLearning}						&   7.01 $\pm$ 0.29                      & 5.73 $\pm$ 0.16                           \\
Mean Teacher \cite{TarvainenMeanResults}  							  &   5.45 $\pm$ 0.14                         &       5.21 $\pm$ 0.21                     \\
VAT (large)    \cite{Miyato2017VirtualLearning} 		  				&                            & 5.77                           \\
VAT+EntMin(Large)\cite{Miyato2017VirtualLearning}   		 &                            & 4.28                           \\ \hline
Improved GAN\cite{SalimansImprovedGANs}     					&     18.44 $\pm$ 4.80                 & 8.11 $\pm$ 1.3                           \\
Improved semi-GAN\cite{Kumar2017ImprovedInvariances} &  4.87 $\pm$1.6                          & 4.39 $\pm$ 1.5   \\
ALI \cite{Dumoulin2017AdversariallyInference}               	  &           					& 7.41 $\pm$ 0.65  \\
Triple-GAN \cite{Li2017TripleNets}                                              &   &       5.77 $\pm$ 0.17\\
Bad GAN \cite{DaiGoodGAN}                	 &                            & 7.42 $\pm$ 0.65               \\
Local GAN \cite{Qi2018GlobalNets}													& 5.48 $\pm$ 0.29         &                   4.73 $\pm$ 0.29 \\   \hline
Improved GAN (ours)                          &  6.13 $\pm$ 0.41                                                                    & 5.6 $\pm$ 0.10                           \\
\textbf{Manifold  regularization (ours)}   &  \textbf{5.67 $\pm$ 0.11}                            &      \textbf{4.63 $\pm$ 0.11}      \\ \hline             
\end{tabular}
\end{table}

We present results on CIFAR-10 \cite{Krizhevsky2009LearningImages} and SVHN \cite{Netzer2011ReadingLearning} in Table \ref{cifar10:sslgan} and Table \ref{svhn:sslgan} respectively. We first note that our implementation of the feature-matching GAN with weight normalization \cite{Salimans2016WeightNetworks} (Improved GAN) \cite{SalimansImprovedGANs} significantly outperforms the original after we tuned training hyperparameters, illustrating the sensitivity of semi-supervised GANs to hyperparameter settings.  

Adding manifold regularization to the feature-matching GAN further improves performance, achieving state-of-the-art results amongst all GAN-based methods, as well as being highly competitive with other non-GAN-based methods. 
We also observe that while simple (isotropic) ambient regularization provides some benefit, our (anisotropic) manifold regularization term provides additional performance gains. 
Our results are consistent with recent work in semi-supervised learning and more generally regularization of neural networks. 
While studies have shown that promoting classifier robustness against local perturbations is effective for semi-supervised learning \cite{Laine2017TemporalLearning,TarvainenMeanResults}, other recent work suggests that it is difficult to achieve local isotropy by enforcing invariance to random perturbations independent of the inputs (which is what the simple ambient regularizer does) in highly non-linear models \cite{adversarial}, so data-dependent perturbations should be used instead. One possibility is to enforce invariance to perturbations along adversarial directions of the classifier \cite{Miyato2017VirtualLearning}. Our approach instead enforces invariance to perturbations on the data manifold as modelled by the GAN.

\subsection{Interaction of our regularizer with the generator}

In the semi-supervised GAN framework, applying manifold regularization to the discriminator has the potential to affect the generator through the adversarial training procedure. We explored the effects of our regularization on the generator by evaluating the quality of images generated using the Inception \cite{SalimansImprovedGANs} and FID scores \cite{Heusel2017GANsEquilibrium} as shown in Table \ref{ics}. We observe that adding manifold regularization yields significant improvements in image quality across both CIFAR-10 and SVHN datasets and with varying amounts of labeled data. These results are consistent with recent work suggesting that gradient penalties on the discriminator may be used to stabilize GAN training \cite{Gulrajani2017ImprovedGANs,MeschederWhichConverge}; our regularization term is closely related to the proposed penalties in \cite{MeschederWhichConverge}. 


\begin{table}
\centering
\caption{Comparison of the Inception \cite{SalimansImprovedGANs} and FID scores\cite{Heusel2017GANsEquilibrium} of our models. Results shown are from 3 runs using different random seeds. }
\begin{tabular}{ll}
\label{ics}
\begin{tabular}{ll}        
\textbf{ CIFAR-10}                       &     Inception Score            \\ \hline
 Unsupervised DCGANs \cite{Radford2016UnsupervisedNetworks}&   6.16 $\pm$ 0.07       \\
 Supervised DCGANs \cite{Radford2016UnsupervisedNetworks} & 6.58  \\
 \begin{tabular}[c]{@{}l@{}}Improved GAN \cite{SalimansImprovedGANs}\\ (minibatch discrimination)\end{tabular}& 8.09 $\pm$ 0.07  \\
 Unsupervised GP-WGAN \cite{Gulrajani2017ImprovedGANs} & 7.86 $\pm$ 0.07 \\ 
Supervised GP-WGAN \cite{Gulrajani2017ImprovedGANs} &8.42 $\pm$ 0.1  \\\hline
\textbf{1000 labels}  &\\
Improved GAN    & 6.28 $\pm$ 0.01      \\
+ Manifold & 6.77 $\pm$ 0.11           \\
\textbf{2000 labels}  &\\
Improved GAN    & 6.24 $\pm$ 0.10       \\
+ Manifold & 6.69 $\pm$ 0.05           \\
\textbf{4000 labels}  &\\
 Improved GAN & 6.24 $\pm$ 0.13                     \\
+ Manifold  & 6.63 $\pm$ 0.09          \\
\end{tabular}
&
\begin{tabular}{ll}
\textbf{CIFAR-10}                           & FID score               \\ \hline
\textbf{1000 labels}  &\\
Improved GAN       &  38.59 $\pm$ 0.18        \\
+ Manifold    &32.03 $\pm$ 0.44          \\
\textbf{2000 labels}  &\\
Improved GAN     &  39.18 $\pm$ 0.62       \\
+ Manifold &33.09 $\pm$ 0.65          \\
\textbf{4000 labels}  &\\
Improved GAN& 39.23       \\
+ Manifold    &33.84 $\pm$ 1.08          \\
\textbf{SVHN}                           &                \\ \hline
\textbf{1000 labels}  &\\
Improved GAN&    86 $\pm$ 12.98        \\
+ Manifold &        90.26 $\pm$ 7    \\
\textbf{500 labels}  &\\
Improved GAN &          85.49 $\pm$ 11.73    \\
+ Manifold &          38.65 $\pm$ 7.33  
\end{tabular}
\end{tabular}
\end{table}

\subsection{Incorporating manifold regularization into convolutional neural nets}

We also explored the potential for our manifold regularization framework to improve performance of classifiers outside the semi-supervised GAN framework. To this end, we performed a series of experiments where we first trained a GAN to learn the marginal distribution of the data $\mathcal{P}_X$  and the data manifold $\mathcal{M}$, and subsequently used the trained GAN to regularize a separate neural network classifier. In this setup, we minimize the following loss where $V$ is the cross entropy loss on the labeled examples
\[L = \frac{1}{n} \sum_n V(x^{(i)},y^{(i)},f) + \gamma_m \Omega(f). \]
Here unlabeled examples are implicitly used to regularize the classifier $f$ since they are used to train the GAN. This setting also enables us to understand how the quality of the generator used to approximate the manifold regularizer affects classification accuracy as there is no interaction between the generator and the classifier being regularized.

We first verified that GANs are able to both learn the data manifold and the density on this manifold on a series of toy examples (Figure 2). In these experiments, we penalized a neural network classifier with our manifold regularizer as approximated with a consensus GAN \cite{Mescheder2017TheGANs}. The classifier we used consists of 6 fully-connected layers with 384 neurons each.
We show further results on variants of these datasets in the Supplementary Material. 
\begin{figure}[h]
\centering
\begin{tabular}{c}
 \includegraphics[width=0.6\linewidth]{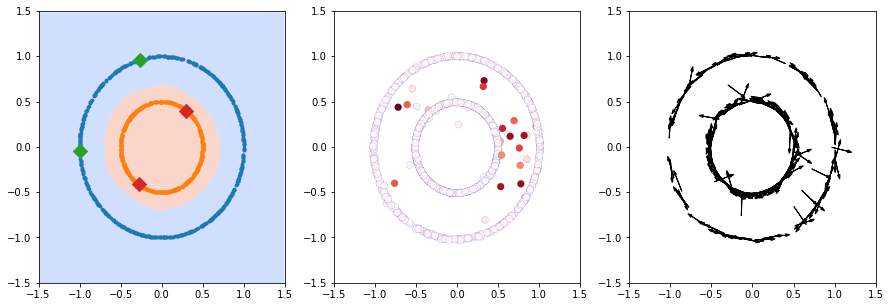}  \\
 \includegraphics[width=0.6\linewidth]{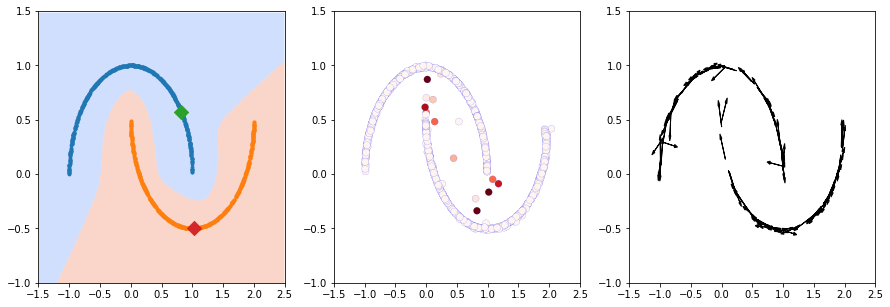}  \\

 \\
\end{tabular}
\caption{Behavior of manifold regularization with a separate classifier on toy examples. Left: Classification boundaries of trained classifier. Labeled examples are shown as points, unlabeled examples were drawn from distributions over the curves shown. The classifier achieves perfect accuracy on the two datasets. 
Middle: Magnitude of the regularization term for a batch of generated samples. Darker fill color reflects larger magnitude. Generated data-points near the decision boundary are highly penalized.
Right: Direction of invariance promoted by our norm. The trained GANs are able to approximate the data distribution and manifold gradients. In this example $\gamma$ = 6 and $\epsilon$  = 0.15.}
\end{figure}

We then evaluated our manifold regularization method on a real image dataset (CIFAR-10), using a standard DCGAN to approximate the manifold regularizer and a 13 layer CNN as the classifier (see Supplementary Material for details). 
We observe that manifold regularization is able to reduce classification error by 2-3\% over the purely supervised baseline across different amounts of labelled data (Table \ref{decoupled-cifar} bottom; first two rows).


In order to quantify the importance of the first step of manifold learning, we also compared the performance of classifiers when regularized using GANs of differing quality (as assessed by the quality of generated images). Specifically, we compare the following generators:
\begin{itemize}
\item A DCGAN which produces decent looking images (inception score of 6.68 and FID of 32.32 on CIFAR-10).
\item A DCGAN with a much lower inception score (inception score of 3.57 on CIFAR-10).
\item A noise image generator where pixels are generated independently from uniform distributions over integers in the range 0 to 255.
\end{itemize}

We observe that using GANs with better generators for manifold regularization resulted in lower classification errors (Table \ref{decoupled-cifar}; last 3 rows). Our results also suggest that even GANs that generate lower quality images but nonetheless have captured some aspects of the image manifold are able to provide some performance benefit when used for manifold regularization. As a negative control, we observe that performance degrades relative to the supervised baseline when we compute the regularizer using randomly generated images.

\begin{table}[h]
\centering
\caption{
Error rate on 4 runs of a CNN with manifold regularization using a separate GAN (DCGAN). Results shown are obtained without data augmentation. Our models were not trained with ZCA whitening but results from other papers include ZCA whitening.}

\label{decoupled-cifar}
\begin{tabular}{lllll}
 																													& 1000 labels (2\%)  &2000 labels (4\%)        & 4000 labels  (8\%)    \\
  \textbf{CIFAR-10}         																			 & 50000 images                   &50000 images   & 50000 images                    \\ \hline
Supervised-only \cite{TarvainenMeanResults}		            &  48.38 $\pm$ 1.07    &  36.07 $\pm$ 0.90     &       24.47 $\pm$ 0.50                        \\
$\Pi$ model \cite{Laine2017TemporalLearning}						&      &    &      16.55$\pm$ 0.29                      \\
Mean Teacher \cite{TarvainenMeanResults}		                 	&   30.62 $\pm$1.13   & 23.14 $\pm$ 0.46   &       17.74 $\pm$ 0.30                      \\  \hline
Supervised-only (ours)																	&  41.65 $\pm$ 3.12    &  32.46 $\pm$ 0.52     &       25.01 $\pm$ 1.29                        \\ 
Supervised + manifold (inception:6.68)  							  &  38.76 $\pm$ 1.81   &  29.44 $\pm$ 0.45    &       23.5 $\pm$ 1.20                 \\ 
Supervised + manifold (inception:3.57)  						       &  39.14 $\pm$ 1.46  & 32.84 $\pm$ 2.13   &       24.87 $\pm$ 0.73               \\
Supervised + manifold (noise)  												   &  66.99 $\pm$ 5.23    & 72.21 $\pm$ 4.87&       67.79 $\pm$ 4.30             \\ \hline
\end{tabular}
\end{table}


\subsection{Understanding our manifold regularization approximation method}

\begin{figure}[h]
    \centering
    \includegraphics[width=1.0\linewidth]{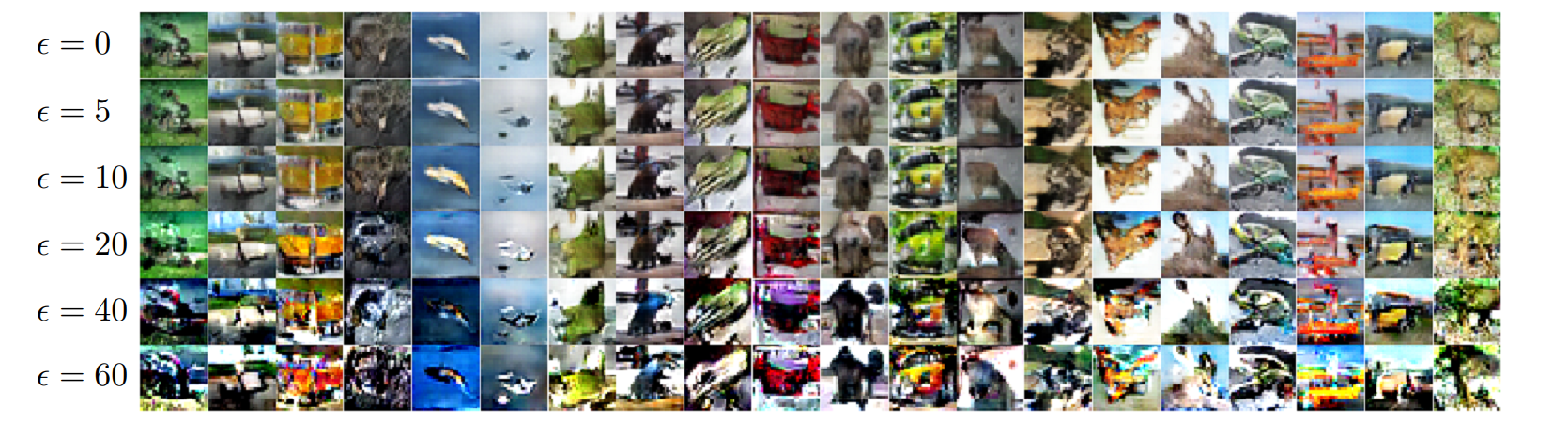}
    \caption{Effect of hyperparameter $\epsilon$ for approximating manifold regularization. Generated images with varying perturbations as per the gradient approximation ($g(z) + \epsilon \bar{r}$) are shown for $\eta=1$. We used $(\epsilon=20,\eta=1)$ in our experiments.}
    \label{tangent}
    \label{2d}
\end{figure}


Finally,  we attempt to provide some intuition of how our manifold regularization approximation method works. Our method promotes classifier invariance between generated samples $g(z)$ and their corresponding perturbations $g(z) + \epsilon \bar{r} $ obtained by perturbing their latent code. In other words, the regularizer promotes invariance of the classifier specifically along directions on the data manifold. We show in Figure \ref{tangent} some examples of generated samples and their corresponding perturbations. We see that even if some features such as the background color may change, there exists a global class consistency between the images as $\epsilon$ is varied. Hence, enforcing invariance in predictions along these directions will result in correct classifications provided $\epsilon$ is not too large (which would result in larger distortions and unrecognizable images). 
An intuitive explanation is that manifold regularization performs label propagation across ``semantically similar'' images by minimizing the manifold consistency cost induced by $\Omega(f)$, such that an image which looks like a red car gets the same label as an orange car.

\section{Conclusion}
GANs are powerful generative models that are able to model the distribution and manifold over natural images. We leverage these properties to perform manifold regularization by approximating a variant of the Laplacian norm using a Monte Carlo approximation that is easily computed with the GAN. We show that our regularization strategy consistently improves classification performance using unlabeled data on the CIFAR-10 and SVHN benchmarks, on several neural network architectures, and with varying amounts of labelled data. In particular, when incorporated into the feature-matching GAN of \cite{SalimansImprovedGANs}, we achieve state-of-the-art results for semi-supervised image classification with a method that is significantly easier to implement than competing methods. We explored the interaction between our regularization and the generator in this framework and reveal a potential connection with gradient penalties for stabilizing GAN training. Using an experimental setup where we decoupled the GAN used for estimating the regularizer and the classifier, we further observed a positive correlation between generator image quality and prediction accuracy. Our work uses GANs in a novel way for semi-supervised classification, and we expect that  our approach will be applicable to semi-supervised regression  \cite{Belkin2004,regressionAnalysis,moscovich2016minimax} as well as unsupervised learning \cite{Belkin2006ManifoldExamples}. 

\section{Acknowlegments}

The computational work for this article was partially performed on resources of the National Supercomputing Centre, Singapore (\url{https://www.nscc.sg}).




\bibliography{nips_2018}

\begin{thebibliography}{30}
\providecommand{\natexlab}[1]{#1}
\providecommand{\url}[1]{\texttt{#1}}
\expandafter\ifx\csname urlstyle\endcsname\relax
  \providecommand{\doi}[1]{doi: #1}\else
  \providecommand{\doi}{doi: \begingroup \urlstyle{rm}\Url}\fi

\bibitem[Belkin and Niyogi(2004)]{Belkin2004}
Mikhail Belkin and Partha Niyogi.
\newblock Semi-supervised learning on riemannian manifolds.
\newblock \emph{Machine Learning}, 56:\penalty0 209--239, 2004.
\newblock ISSN 1573-0565.
\newblock \doi{10.1023/B:MACH.0000033120.25363.1e}.
\newblock URL \url{https://doi.org/10.1023/B:MACH.0000033120.25363.1e}.

\bibitem[Belkin et~al.(2006)Belkin, Niyogi, and
  Sindhwani]{Belkin2006ManifoldExamples}
Mikhail Belkin, Partha Niyogi, and Vikas Sindhwani.
\newblock {Manifold Regularization: A Geometric Framework for Learning from
  Labeled and Unlabeled Examples}.
\newblock \emph{Journal of Machine Learning Research}, 2006.
\newblock URL \url{http://www.jmlr.org/papers/volume7/belkin06a/belkin06a.pdf}.

\bibitem[Bousquet et~al.(2003)Bousquet, Chapelle, and
  Hein]{BousquetMeasureRegularization}
Olivier Bousquet, Olivier Chapelle, and Matthias Hein.
\newblock {Measure Based Regularization}.
\newblock \emph{NIPS}, 2003.
\newblock URL
  \url{https://pdfs.semanticscholar.org/cbe5/8390608078c3187e34f202c67d50044ce7d6.pdf}.

\bibitem[Chapelle et~al.(2010)Chapelle, Schlkopf, and
  Zien]{ChapelleSemi-SupervisedBy}
Olivier Chapelle, Bernhard Schlkopf, and Alexander Zien.
\newblock \emph{Semi-Supervised Learning}.
\newblock The MIT Press, 1st edition, 2010.
\newblock ISBN 0262514125, 9780262514125.

\bibitem[Dai et~al.(2017)Dai, Yang, Yang, Cohen, and Salakhutdinov]{DaiGoodGAN}
Zihang Dai, Zhilin Yang, Fan Yang, William~W Cohen, and Ruslan Salakhutdinov.
\newblock {Good Semi-supervised Learning That Requires a Bad GAN}.
\newblock \emph{NIPS}, 2017.
\newblock URL \url{https://arxiv.org/pdf/1705.09783.pdf}.

\bibitem[Dumoulin et~al.(2017)Dumoulin, Belghazi, Poole, Mastropietro, Lamb,
  Arjovsky, and Courville]{Dumoulin2017AdversariallyInference}
Vincent Dumoulin, Ishmael Belghazi, Ben Poole, Olivier Mastropietro, Alex Lamb,
  Martin Arjovsky, and Aaron Courville.
\newblock {Adversarially Learned Inference}.
\newblock \emph{ICLR}, 2017.
\newblock URL \url{http://arxiv.org/abs/1606.00704}.

\bibitem[Gulrajani et~al.(2017)Gulrajani, Ahmed, Arjovsky, Dumoulin, and
  Courville]{Gulrajani2017ImprovedGANs}
Ishaan Gulrajani, Faruk Ahmed, Martin Arjovsky, Vincent Dumoulin, and Aaron
  Courville.
\newblock {Improved Training of Wasserstein GANs}.
\newblock \emph{NIPS}, 2017.
\newblock URL \url{http://arxiv.org/abs/1704.00028}.

\bibitem[Heusel et~al.(2017)Heusel, Ramsauer, Unterthiner, Nessler, and
  Hochreiter]{Heusel2017GANsEquilibrium}
Martin Heusel, Hubert Ramsauer, Thomas Unterthiner, Bernhard Nessler, and Sepp
  Hochreiter.
\newblock {GANs Trained by a Two Time-Scale Update Rule Converge to a Local
  Nash Equilibrium}.
\newblock \emph{NIPS}, 2017.
\newblock URL \url{https://arxiv.org/pdf/1706.08500.pdf}.

\bibitem[Krizhevsky(2009)]{Krizhevsky2009LearningImages}
Alex Krizhevsky.
\newblock {Learning Multiple Layers of Features from Tiny Images}.
\newblock Technical report, 2009.

\bibitem[Kumar et~al.(2017)Kumar, Sattigeri, and
  Fletcher]{Kumar2017ImprovedInvariances}
Abhishek Kumar, Prasanna Sattigeri, and P.~Thomas Fletcher.
\newblock {Improved Semi-supervised Learning with GANs using Manifold
  Invariances}.
\newblock \emph{NIPS}, 2017.
\newblock URL \url{http://arxiv.org/abs/1710.09829}.

\bibitem[Lafferty and Wasserman(2007)]{regressionAnalysis}
John Lafferty and Larry Wasserman.
\newblock {Statistical Analysis of Semi-Supervised Regression}.
\newblock \emph{NIPS}, 2007.

\bibitem[Laine and Aila(2017)]{Laine2017TemporalLearning}
Samuli Laine and Timo Aila.
\newblock {Temporal Ensembling for Semi-Supervised Learning}.
\newblock \emph{ICLR}, 2017.
\newblock URL \url{http://arxiv.org/abs/1610.02242}.

\bibitem[Li et~al.(2017)Li, Xu, Zhu, and Zhang]{Li2017TripleNets}
Chongxuan Li, Kun Xu, Jun Zhu, and Bo~Zhang.
\newblock {Triple Generative Adversarial Nets}.
\newblock \emph{NIPS}, 2017.
\newblock URL \url{http://arxiv.org/abs/1703.02291}.

\bibitem[Mescheder et~al.(2017)Mescheder, Nowozin, and
  Geiger]{Mescheder2017TheGANs}
Lars Mescheder, Sebastian Nowozin, and Andreas Geiger.
\newblock {The Numerics of GANs}.
\newblock \emph{NIPS}, 2017.
\newblock URL \url{https://arxiv.org/pdf/1705.10461.pdf}.

\bibitem[Mescheder et~al.(2018)Mescheder, Geiger, and
  Nowozin]{MeschederWhichConverge}
Lars Mescheder, Andreas Geiger, and Sebastian Nowozin.
\newblock {Which Training Methods for GANs do actually Converge}.
\newblock \emph{arXiv preprint arXiv:1801.04406}, 2018.

\bibitem[Miyato et~al.(2017)Miyato, Maeda, Koyama, and
  Ishii]{Miyato2017VirtualLearning}
Takeru Miyato, Shin-ichi Maeda, Masanori Koyama, and Shin Ishii.
\newblock {Virtual Adversarial Training: a Regularization Method for Supervised
  and Semi-supervised Learning}.
\newblock \emph{arXiv preprint arXiv:1704.03976}, 2017.
\newblock URL \url{http://arxiv.org/abs/1704.03976}.

\bibitem[Moscovich et~al.(2017)Moscovich, Jaffe, and
  Nadler]{moscovich2016minimax}
Amit Moscovich, Ariel Jaffe, and Boaz Nadler.
\newblock Minimax-optimal semi-supervised regression on unknown manifolds.
\newblock \emph{AISTATS}, 2017.

\bibitem[Netzer et~al.(2011)Netzer, Wang, Coates, Bissacco, Wu, and
  Ng]{Netzer2011ReadingLearning}
Yuval Netzer, Tao Wang, Adam Coates, Alessandro Bissacco, Bo~Wu, and Andrew~Y
  Ng.
\newblock {Reading Digits in Natural Images with Unsupervised Feature
  Learning}.
\newblock \emph{NIPS}, 2011.
\newblock URL
  \url{http://ufldl.stanford.edu/housenumbers/nips2011_housenumbers.pdf}.

\bibitem[Qi et~al.(2018)Qi, Zhang, Hu, Edraki, Wang, and Hua]{Qi2018GlobalNets}
Guo-Jun Qi, Liheng Zhang, Hao Hu, Marzieh Edraki, Jingdong Wang, and Xian-Sheng
  Hua.
\newblock {Global versus Localized Generative Adversarial Nets}.
\newblock \emph{CVPR}, 2018.
\newblock URL \url{http://arxiv.org/abs/1711.06020}.

\bibitem[Radford et~al.(2016)Radford, Metz, and
  Chintala]{Radford2016UnsupervisedNetworks}
Alec Radford, Luke Metz, and Soumith Chintala.
\newblock {Unsupervised Representation learning with deep convolutional
  generative adversarial networks}.
\newblock \emph{ICLR}, 2016.
\newblock URL \url{https://arxiv.org/pdf/1511.06434.pdf}.

\bibitem[Rifai et~al.(2011{\natexlab{a}})Rifai, Dauphin, Vincent, Bengio, and
  Muller]{Rifai2011TheClassifier}
Salah Rifai, Yann~N. Dauphin, Pascal Vincent, Yoshua Bengio, and Xavier Muller.
\newblock {The Manifold Tangent Classifier}.
\newblock \emph{NIPS}, 2011{\natexlab{a}}.
\newblock URL
  \url{https://papers.nips.cc/paper/4409-the-manifold-tangent-classifier}.

\bibitem[Rifai et~al.(2011{\natexlab{b}})Rifai, Vincent, Muller, Glorot, and
  Bengio]{Rifai2011ContractiveExtraction}
Salah Rifai, Pascal Vincent, Xavier Muller, Xavier Glorot, and Yoshua Bengio.
\newblock {Contractive Auto-Encoders: Explicit Invariance During Feature
  Extraction}.
\newblock \emph{ICML}, 2011{\natexlab{b}}.
\newblock URL \url{http://www.icml-2011.org/papers/455_icmlpaper.pdf}.

\bibitem[Salimans and Kingma(2016)]{Salimans2016WeightNetworks}
Tim Salimans and Diederik~P Kingma.
\newblock {Weight Normalization: A Simple Reparameterization to Accelerate
  Training of Deep Neural Networks}.
\newblock \emph{NIPS}, 2016.
\newblock URL \url{http://arxiv.org/abs/1602.07868}.

\bibitem[Salimans et~al.(2016)Salimans, Goodfellow, Zaremba, Cheung, Radford,
  and Chen]{SalimansImprovedGANs}
Tim Salimans, Ian Goodfellow, Wojciech Zaremba, Vicki Cheung, Alec Radford, and
  Xi~Chen.
\newblock {Improved Techniques for Training GANs}.
\newblock \emph{NIPS}, 2016.
\newblock URL \url{https://arxiv.org/pdf/1606.03498.pdf}.

\bibitem[Shao et~al.(2017)Shao, Kumar, and Fletcher]{Shao2017TheModels}
Hang Shao, Abhishek Kumar, and P.~Thomas Fletcher.
\newblock {The Riemannian Geometry of Deep Generative Models}.
\newblock \emph{arXiv preprint arXiv:1711.08014}, 2017.

\bibitem[Simard et~al.(1991)Simard, Victorri, LeCun, and
  Denker]{Simard1991TangentNetwork}
Patrice Simard, Bernard Victorri, Yann LeCun, and John Denker.
\newblock {Tangent Prop - A formalism for specifying selected invariances in an
  adaptive network}.
\newblock \emph{NIPS}, 1991.
\newblock URL
  \url{https://papers.nips.cc/paper/536-tangent-prop-a-formalism-for-specifying-selected-invariances-in-an-adaptive-network}.

\bibitem[{Springenberg}(2016)]{catgan}
J.~T. {Springenberg}.
\newblock {Unsupervised and Semi-supervised Learning with Categorical
  Generative Adversarial Networks}.
\newblock \emph{ICLR}, 2016.

\bibitem[{Szegedy} et~al.(2014){Szegedy}, {Zaremba}, {Sutskever}, {Bruna},
  {Erhan}, {Goodfellow}, and {Fergus}]{adversarial}
C.~{Szegedy}, W.~{Zaremba}, I.~{Sutskever}, J.~{Bruna}, D.~{Erhan},
  I.~{Goodfellow}, and R.~{Fergus}.
\newblock {Intriguing properties of neural networks}.
\newblock \emph{ICLR}, December 2014.

\bibitem[Tarvainen and Valpola(2017)]{TarvainenMeanResults}
Antti Tarvainen and Harri Valpola.
\newblock {Mean teachers are better role models: Weight-averaged consistency
  targets improve semi-supervised deep learning results}.
\newblock \emph{NIPS}, 2017.
\newblock URL \url{https://arxiv.org/pdf/1703.01780.pdf}.

\bibitem[Zhu et~al.(2016)Zhu, Kr{\"{a}}henb{\"{u}}hl, Shechtman, and
  Efros]{Zhu2016GenerativeManifold}
Jun-Yan Zhu, Philipp Kr{\"{a}}henb{\"{u}}hl, Eli Shechtman, and Alexei~A.
  Efros.
\newblock {Generative Visual Manipulation on the Natural Image Manifold}.
\newblock \emph{ECCV}, 2016.
\newblock URL \url{http://arxiv.org/abs/1609.03552}.

\end{thebibliography}
\bibliographystyle{plain}

\end{document}


\appendix
\addcontentsline{toc}{section}{Appendices}
\section*{Appendices}

\section{Semi supervised learning results }
\begin{table}[h]
\centering
\caption{Error rate on CIFAR-10 average over 4 runs with different random seeds. Results were obtained without data augmentation. We also reported performance of our model using our norm approximated with the standard finite stochastic difference method.}
\label{cifar10:sslgan}
\begin{tabular}{lllll}
                          										 									& 1000 labels (2\%)  &2000 labels (4\%)        & 4000 labels  (8\%)          \\
  \textbf{CIFAR-10}         																			 & 50000 images                   &50000 images   & 50000 images                    \\ \hline

$\Pi$ model 					&     &    &       16.55 $\pm$ 0.29                      \\
Mean Teacher 		                 	&   30.62 $\pm$1.13   & 23.14 $\pm$ 0.46   &       17.74 $\pm$ 0.30                      \\  

VAT (large)                    &                                    & & 14.18                      \\
VAT+EntMin(Large)       &                                   &   & 13.15                        \\ \hline
Improved GAN       	             & 21.83 $\pm$ 2.01           &19.61 $\pm$ 2.09    & 18.63 $\pm$ 2.32            \\
Improved Semi-GAN &  19.52 $\pm$1.5    &    & 16.20 $\pm$ 1.6                \\
ALI                  & 19.98 $\pm$ 0.89           &19.09 $\pm$ 0.44   & 17.99 $\pm$ 1.62  \\
Triple-GAN                                             &									&							&16.99 $\pm$ 0.36\\
Bad GAN                            			             &                                   &                            & 14.41 $\pm$ 0.30              \\
Local GAN												& 17.44 $\pm$ 0.25       &&                   14.23 $\pm$ 0.27  \\   \hline
Improved GAN (ours)                                                                    &17.50 $\pm$ 0.34       &    16.80 $\pm$ 0.07   & 15.5 $\pm$ 0.35    \\
Ambient regularization (ours)                                    	&16.81 $\pm$ 0.21                       & 15.99 $\pm$ 0.14                                             &  14.75 $\pm$ 0.37                         \\
Manifold reg. 1st method         &                           &                                             &    14.40$\pm$0.21                          \\  
\textbf{Manifold  regularization (ours)}                                                & \textbf{16.37 $\pm$ 0.42}  & \textbf{15.25 $\pm$ 0.35}                             &\textbf{14.34 $\pm$ 0.17}                  \\ \hline
\end{tabular}
\end{table}

\begin{table}[h]
\centering
\caption{Error rate on SVHN average over 4 runs with different random seeds. Results were obtained without data augmentation.We also reported performance of our model using the our norm approximated with the standard finite stochastic difference method.}
\label{svhn:sslgan}
\begin{tabular}{lllll}
                                                                & 500 labels(0.3\%)       & 1000 labels(1.4\%)    \\
 \textbf{SVHN}                                       							 & 73257 images & 73257 images \\ \hline
$\Pi$ model 						&   7.01 $\pm$ 0.29                      & 5.73 $\pm$ 0.16                           \\
Mean Teacher 							  &   5.45 $\pm$ 0.14                         &       5.21 $\pm$ 0.21                     \\
VAT (large)   		  				&                            & 5.77                           \\
VAT+EntMin(Large) 		 &                            & 4.28                           \\ \hline
Improved GAN    					&     18.44 $\pm$ 4.80                 & 8.11 $\pm$ 1.3                           \\
Improved semi-GAN &  4.87 $\pm$1.6                          & 4.39 $\pm$ 1.5   \\
ALI               	  &           					& 7.41 $\pm$ 0.65  \\
Triple-GAN                                              &   &       5.77 $\pm$ 0.17\\
Bad GAN               	 &                            & 7.42 $\pm$ 0.65               \\
Local GAN													& 5.48 $\pm$ 0.29         &                   4.73 $\pm$ 0.29 \\   \hline
Improved GAN (ours)                          &  6.13 $\pm$ 0.41                                                                    & 5.6 $\pm$ 0.10                           \\
Manifold reg. 1st method     &                            &     \textbf{4.51 $\pm$ 0.22}              \\ 

\textbf{Manifold  regularization (ours)}   &  \textbf{5.67 $\pm$ 0.11}                            &      \textbf{4.63 $\pm$ 0.11}      \\ \hline             
\end{tabular}
\end{table}

\section{Architectures and hyperparameters}

\subsection{Manifold regularization on semi-supervised feature-matching GANs}

\begin{table}[H]
\centering
\caption{Generator architecture we used for our semi-supervised GAN experiments.}
\label{archi2}
\begin{tabular}{c}
\hline
CIFAR-10 \& SVHN                    \\ \hline
latent space 100 (uniform noise)                            \\
dense 4 $\times$ 4 $\times$ 512 batchnorm ReLU  \\
5$\times$5 conv.T stride=2 256 batchnorm ReLU \\
5$\times$5 conv.T stride=2 128 batchnorm ReLU \\
5$\times$5 conv.T stride=2 3 weightnorm tanh \\ \hline
\end{tabular}
\end{table}

\begin{table}[H]
\centering
\caption{Discriminator architecture we used in our semi-supervised GAN experiments.}
\label{archi1}
\begin{tabular}{cc}
\hline
conv-large CIFAR-10                                 & conv-small SVHN                \\ \hline
\multicolumn{2}{c}{32$\times$32$\times$3 RGB images}                                                                       \\
\multicolumn{2}{c}{dropout, $p=0.2$}                                                                             \\ \hline
\multicolumn{1}{c|}{3$\times$3 conv. weightnorm 96 lReLU}  & 3$\times$3 conv. weightnorm 64 lReLU  \\
\multicolumn{1}{c|}{3$\times$3 conv. weightnorm 96 lReLU}  & 3$\times$3 conv. weightnorm 64 lReLU  \\
\multicolumn{1}{c|}{3$\times$3 conv. weightnorm 96 lReLU stride=2}  & 3$\times$3 conv. weightnorm 64 lReLU stride=2  \\ \hline
\multicolumn{2}{c}{dropout, $p=0.5$}                                                                             \\ \hline
\multicolumn{1}{c|}{3$\times$3 conv. weightnorm 192 lReLU} & 3$\times$3 conv. weightnorm 128 lReLU \\
\multicolumn{1}{c|}{3$\times$3 conv. weightnorm 192 lReLU} & 3$\times$3 conv. weightnorm 128 lReLU \\
\multicolumn{1}{c|}{3$\times$3 conv. weightnorm 192 lReLU stride=2} & 3$\times$3 conv. weightnorm 128 lReLU stride=2 \\ \hline
\multicolumn{2}{c}{dropout, $p=0.5$}                                                                             \\ \hline
\multicolumn{1}{c|}{3$\times$3 conv. weightnorm 192 lReLU pad=0} & 3$\times$3 conv. weightnorm 128 lReLU pad=0 \\
\multicolumn{1}{c|}{NiN weightnorm 192 lReLU}                     & NiN weightnorm 128 lReLU                     \\
\multicolumn{1}{c|}{NiN weightnorm 192 lReLU}                     & NiN weightnorm 128 lReLU                     \\ \hline
\multicolumn{2}{c}{global-pool}                                                                                  \\
\multicolumn{2}{c}{dense weightnorm 10}                                                                          \\ \hline
\end{tabular}
\end{table}

\begin{table}[H]
\centering
\caption{Hyperparameters of the models in our semi-supervised GAN experiments.}
\label{my-label}
\begin{tabular}{lll}
Hyperparameters       & CIFAR                                         & SVHN            \\ \hline
$\gamma$              & $10^{-3}$                                     & $10^{-3}$       \\
$\epsilon$            & $20$                                          & $20$            \\
$\eta$                & $1$                                           & $1$             \\
Epoch                 & 1400                                          & 400             \\
Batch size            & 25                                            & 50              \\
Optimizer             & \multicolumn{2}{l}{ADAM($\alpha=3*10^{-4}, beta1=0.5)$}         \\
Learning rate         & linearly decayed to 0 after 1200 epochs       & no decay        \\
Leaky ReLU slope      & \multicolumn{2}{l}{0.2}                                         \\
Weight initialization & \multicolumn{2}{l}{Isotropic gaussian ($\mu = 0, \sigma=0.05$)} \\
Biais initialization  & \multicolumn{2}{l}{Constant(0)}                                 \\ \hline
\end{tabular}
\end{table}

\subsection{Manifold regularization on convolutional neural nets}

\begin{table}[H]
\centering
\caption{The convolutional neural network architecture we used for experiments in Section 4.4.}
\label{my-label}
\begin{tabular}{c}
CIFAR-10 \& SVHN convnet                     \\ \hline
32 $\times$ 32 $\times$ 3                     \\
3 $\times$ 3, 96 conv. batchnorm lReLU        \\
3 $\times$ 3, 96 conv. batchnorm lReLU        \\
3 $\times$ 3, 96 conv. batchnorm lReLU        \\ \hline
2 $\times$ 2 maxpool                          \\
dropout = 0.5                                 \\ \hline
3 $\times$ 3, 128 conv. batchnorm lReLU \\
3 $\times$ 3, 128 conv. batchnorm lReLU       \\
3 $\times$ 3, 128 conv. batchnorm lReLU       \\ \hline
2 $\times$ 2 maxpool                          \\
dropout=0.5                                   \\ \hline
3 $\times$ 3, 256 conv. batchnorm lReLU  pad=0    \\
1 $\times$ 1, 128 conv. batchnorm lReLU       \\
1 $\times$ 1, 128 conv. batchnorm lReLU       \\ \hline
average pooling                               \\
10 dense                                      \\ \hline
\end{tabular}
\end{table}

\begin{table}[H]
\centering
\caption{Hyperparameters of the models used in Section 4.4.}
\label{my-label}
\begin{tabular}{ll}
Hyperparameters       & CIFAR                                          \\ \hline
$\gamma$              & $10^{-4}$                                     \\
$\epsilon$            & $20$                                             \\
$\eta$                & $1$                                              \\
Epoch                 & 200                                             \\
Batch size            &100                                             \\
Optimizer             & ADAM($\alpha=3*10^{-4}, beta1=0.9)$        \\
Learning rate         & no decay             \\
Leaky ReLU slope      & 0.2                                         \\
Weight initialization & Isotropic gaussian ($\mu = 0, \sigma=0.05$) \\
Biais initialization  & Constant(0)                                 \\ \hline
\end{tabular}
\end{table}

\begin{table}[H]
\centering
\caption{GAN architecture we used for our experiments in Section 4.4.}
\label{my-label}
\begin{tabular}{cc}
Discriminator                                 & Generator                                     \\ \hline
32$\times$32$\times$3 RGB images              & latent space 100 (gaussian noise)             \\
4$\times$4 64 conv stride=2 batchnorm lReLU   & 4$\times$4 1024 batchnorm conv.T pad=0 ReLU   \\
4$\times$4 256 conv stride=2 batchnorm lReLU  & 4$\times$4 256 stride=2 batchnorm conv.T ReLU \\
4$\times$4 1024 conv stride=2 batchnorm lReLU & 4$\times$4 64 stride=2 batchnorm conv.T ReLU  \\
4$\times$4 1 conv pad=0                       & 4$\times$4 3 stride=2 batchnorm conv.T tanh   \\ \hline
\end{tabular}
\end{table}

\subsection{Toy examples}

For the GAN, we used an simple setup where the generator and discriminator both have 6 fully-connected layers of 384 neurons with ReLU activations. The latent space of the generator has two dimensions. The generator has two outputs and the discriminator has one. The GAN was trained with RMSProp and consensus optimization, which is very effective in stabilizing the training of the GAN and to enable it to capture highly multi modal distributions. //
The neural network we used for the classifier has a similar architecture to the discriminator with 6 fully-connected layers of 384 neurons with ReLU activations. 

\section{Inception scores}


\section{Generated images}
\begin{figure}[H]
    \centering
    \includegraphics[width=0.8\linewidth]{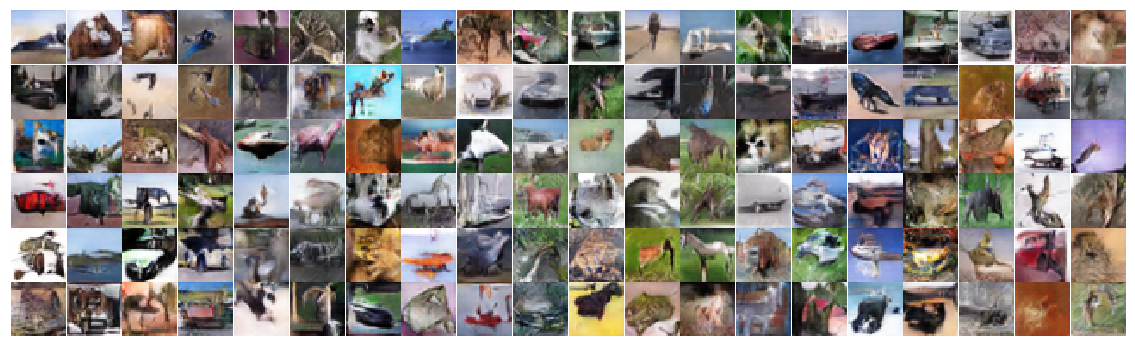}
    \caption{Generated images for GANs used in Section 4.4.}
    \label{2d}
\end{figure}
\begin{figure}[H]
    \centering
    \includegraphics[width=0.8\linewidth]{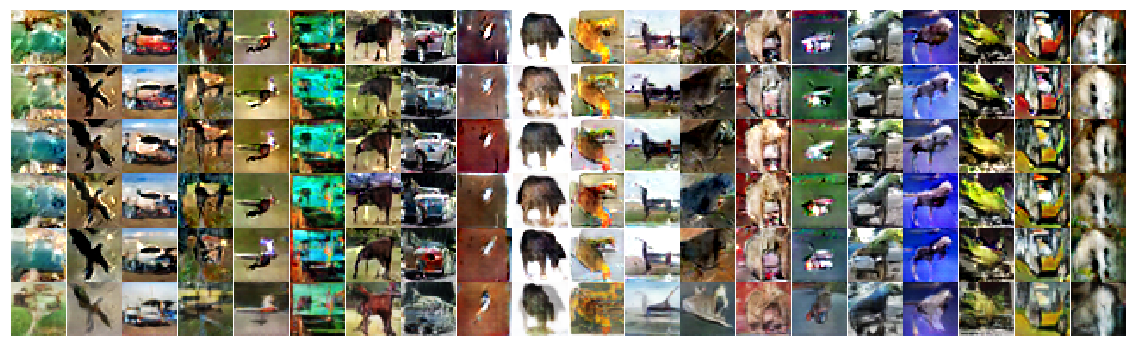}
    \caption{Moving $\eta$ in the range $[10^{-4},10^{-3},10^{-2},10^{-1},1,10]$ with $\epsilon = 20$  fixed }
    \label{2d}
\end{figure}
\begin{figure}[H]
    \centering
    \includegraphics[width=0.8\linewidth]{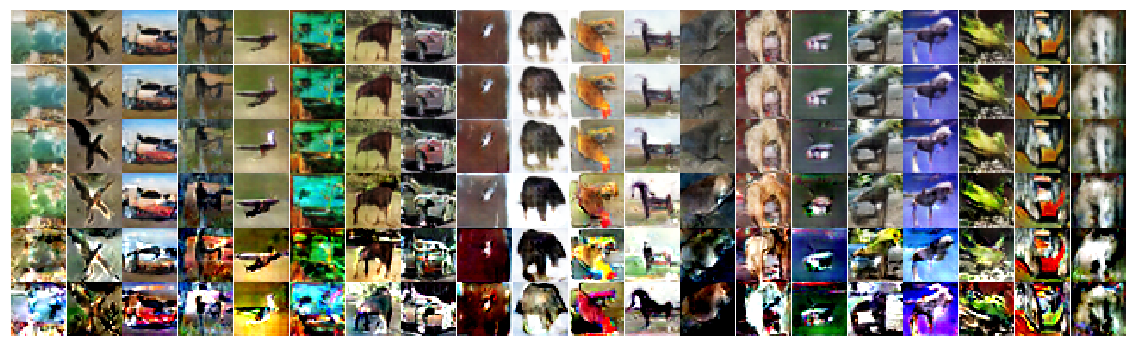}
    \caption{Moving $\epsilon$ in the range $[0,5,10,20,40,60]$ with $\eta= 1$  fixed}
    \label{2d}
\end{figure}

\section{Unsupervised learning of simple manifolds by minimizing our norm}

\begin{figure}[H]
\centering
\begin{tabular}{ccc}
 \includegraphics[width=0.2\linewidth]{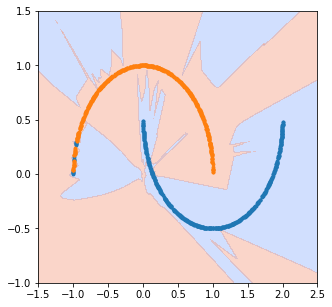}  &  \includegraphics[width=0.2\linewidth]{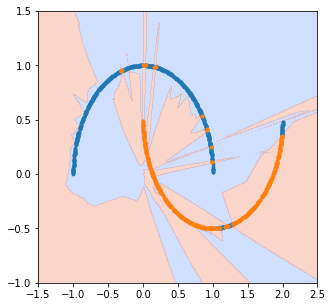}  &  \includegraphics[width=0.2\linewidth]{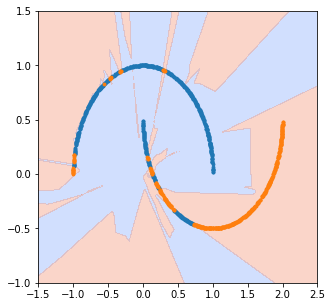} \\
 \includegraphics[width=0.2\linewidth]{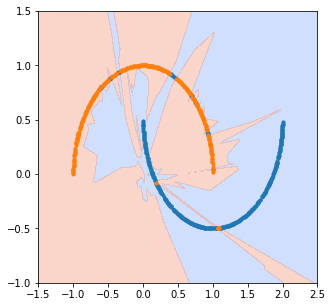}  &  \includegraphics[width=0.2\linewidth]{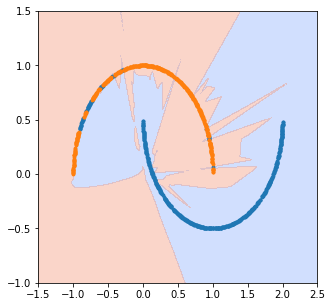}  &  \includegraphics[width=0.2\linewidth]{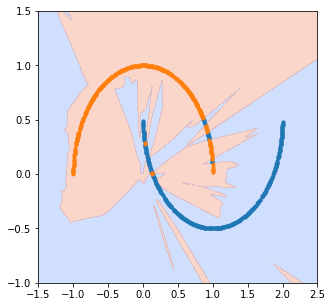} \\ 
\end{tabular}
\caption{Unsupervised learning of the two moon dataset. No labels are used for the training. We are minimizing a loss function including  our norm, ridge penalty and an entropy term $\Omega(f) = \gamma_L \left\| f \right\|_L +\gamma_K \left\| f \right \|_K^2 - \gamma_h H(f) $. We added the entropy to avoid degenerated solutions. In this example $\epsilon=0.15$ , $\gamma=3$ ,$\gamma_K=1$ , $\gamma_h=0.1$  }
\end{figure}

\section{Learning of a noisy toy dataset with a GAN}
\begin{figure}[H]
    \centering
 \includegraphics[width=0.8\linewidth]{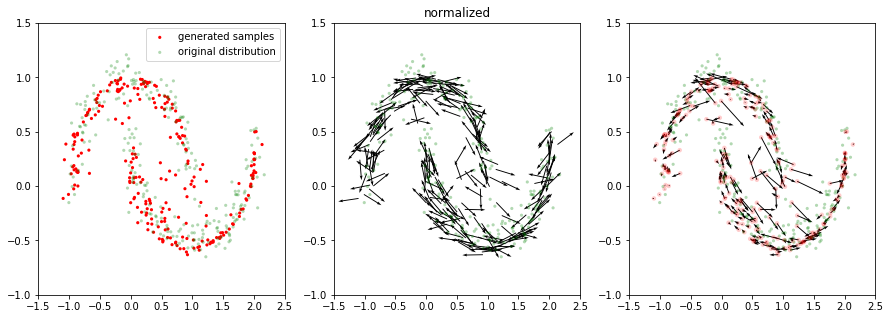} 
    \caption{Unsupervised learning of a noisy manifold by a Consensus GAN.}
    \label{2d}
\end{figure}
\begin{figure}[H]
\centering
\begin{tabular}{ccc}
 \includegraphics[width=0.2\linewidth]{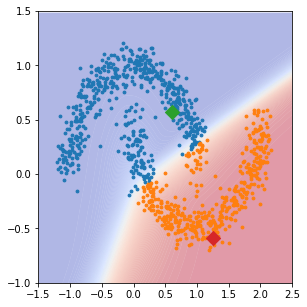}  &  \includegraphics[width=0.2\linewidth]{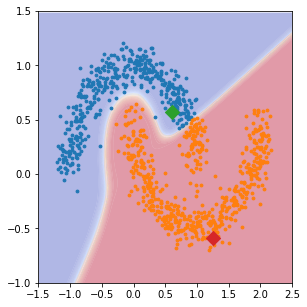}  &  \includegraphics[width=0.2\linewidth]{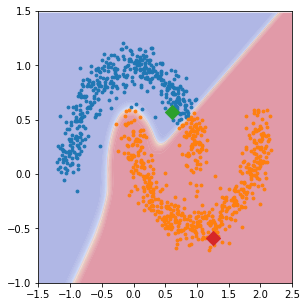} \\
 \includegraphics[width=0.2\linewidth]{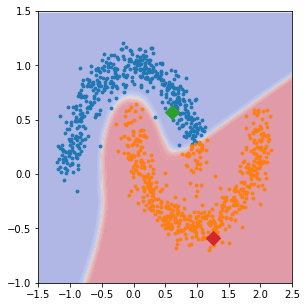}  &  \includegraphics[width=0.2\linewidth]{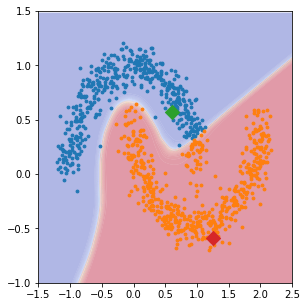}  &  \includegraphics[width=0.2\linewidth]{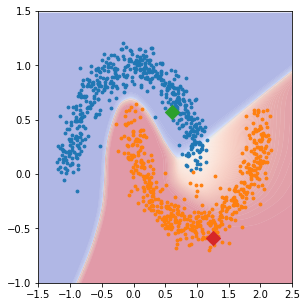} \\ 
  \includegraphics[width=0.2\linewidth]{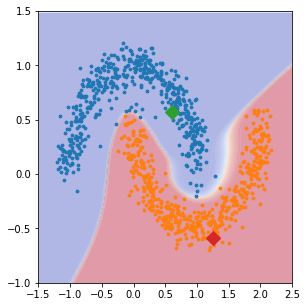}  &  \includegraphics[width=0.2\linewidth]{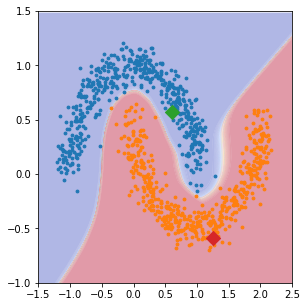}  &  \includegraphics[width=0.2\linewidth]{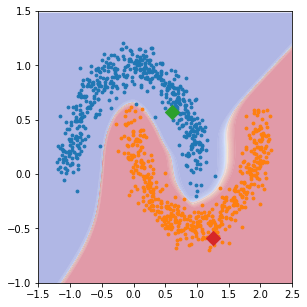} \\ 
\end{tabular}
\caption{Semi-supervised learning of the the noisy two moons dataset after {200,400,800,1000,1200,1400,1600,1800,2000} iterations. We chose for all our 2d experiments $\epsilon=0.15$, $\gamma$ = 6 ,$
\eta=0.01$}
\end{figure}